\documentclass[a4paper]{article}
\pdfoutput=1
\usepackage{dblfloatfix}
\usepackage{caption}  

\usepackage{graphicx}
\usepackage[numbers]{natbib}

\usepackage{tikz}
\usetikzlibrary{shapes,arrows,calc,positioning}

\usepackage{amssymb}
\usepackage{amsmath}

\usepackage{algorithm}
\usepackage{algorithmicx}
\usepackage{algpseudocode}

\usepackage{subfigure}
\usepackage{paralist}

\usepackage{listings}
\usepackage{url}

\algnewcommand{\algorithmicgoto}{\textbf{go to}}%
\algnewcommand{\GoTo}[1]{\algorithmicgoto~\ref{#1}}%

\newcommand{\keywords}[1]{\par\addvspace\baselineskip\noindent\textbf{Keywords: }\textit{#1}.}

\begin{document}
\title{Turing's Imitation Game has been Improved}

\author{Norbert~B\'atfai\thanks{N. B\'atfai is with the Department of Information Technology, University of Debrecen, H-4010 Debrecen PO Box 12, Hungary, e-mail: batfai.norbert@inf.unideb.hu.}}

\maketitle

\begin{abstract}
Using the recently introduced universal computing model, called orchestrated machine, that represents computations in a dissipative environment, we consider a new kind of interpretation of Turing's Imitation Game.
In addition we raise the question whether the intelligence may show fractal properties. Then we sketch a vision of what robotic cars are going to do in the future. Finally we give the specification of an artificial life game based on the concept of orchestrated machines. 
The purpose of this paper is to start the search for possible relationships between these different topics.
\keywords machine intelligence, Turing test, self-similarity dimension, orchestrated machines, driverless car
\end{abstract}

\section{Introduction}

The manuscript 
\cite{OrchMach}
has introduced two intelligence functions $IQ$ and $EQ$ for the simplest computer programs (to be full precise for Turing machines with a given number of states) that try to measure a kind of an a priori  ability of the programs to cooperate among each other. In intuitive sense, the function $EQ(N)$ is the maximum number of machines that can do $N$ steps together on the universal orchestrated machine 
%($\operatorname{OrchMach1}$) 
and inversely, $IQ(Z)$ is the maximum number of steps that can be done by $Z$ machines together.

At the heart of the concept of orchestrated machines is the testing the ability of Turing machines to cooperate each other.  The Turing test also serves a similar aim in sense that it was designed to test the ability of a computer program to communicate with a human, and it may be therefore appropriate to think about the Turing test from the point of view of orchestrated machines.  It will be the main subject of this paper.

The initial estimation of the empirical intelligence functions shown in \cite{OrchMach} 
have already suggested that there may be an interesting correlation between $IQ$ and $EQ$.
Namely, the question could be raised whether the intelligence has fractal nature.
This issue will be raised in Sect. \ref{ref_turingtest}, where in addition  
we sketch a vision of how drivers may change their attitude to their driverless cars. Then finally, Sect. \ref{ref_game} will introduce the specification of a new artificial life game. Through of this paper we use the notations of the previous paper \cite{OrchMach}.
 
\section{Orchestrated machine algorithms as Turing tests}\label{ref_turingtest}

If we observe the common conversations of people we may find in many cases that we cannot determine the intelligence of  the participants, like we cannot tell the size of a cloud if we can see only the cloud itself. 
In a sense, most of the conversations are similar to themselves, in point of fact, common sense of the people is similar to itself.
(Actually, it is not surprising at all that conversations may have fractal properties, because, for example, it has already been known that written corpora or the language itself have fractal properties \cite{textfractals},  \cite{fractlang}.) 
This scale-free simple observation suggests to try to introduce the fractal dimension of conversations that would be definable by using the self-similarity dimension \cite{frakbook} in the form 
$log_{\left \lfloor \overline{o_2} \right \rfloor}N$
or in the more intuitive form
$log_{EQ}IQ$, where $N$ denotes the length of the conversation
and $\left \lfloor \overline{o_2} \right \rfloor$ denotes the number of participants in the conversation. To be more precise, using the terminology of \cite{OrchMach}, $IQ$ is the intelligence
quotient and $EQ$ is the emotional intelligence quotient.

\begin{figure}[t]
\centering
\begin{tikzpicture}[ scale=0.65]
\node[label={[align=center]chat}]  at (-4.25,11) {};
\node (v2) at (-6,4.5) {};
\node (v1) at (-6,12) {};
\node (v3) at (-2.5,12) {};
\node (v4) at (-2.5,4.5) {};
\draw  (v1) edge (v2);
\draw  (v3) edge (v4);
\node (v5) at (-6,10) {};
\node (v6) at (-2.5,10) {};
\node (v12) at (-6,6.5) {};
\node (v11) at (-2.5,6.5) {};
\tikzstyle{myedgestyle} = [-latex]
\draw[-latex]  (v5) edge (v6);
\node (v8) at (-6,9.5) {};
\node (v7) at (-2.5,9.5) {};
\draw[-latex]  (v7) edge (v8);
\node at (-3,10.5) {X};
\node (v9) at (-6,7) {};
\node (v10) at (-2.5,7) {};
\draw[-latex]  (v9) edge (v10);
\draw[-latex]  (v11) edge (v12);
\node at (-3,7.5) {Y};
\node[label={[align=center]below:(simulation/\\interpretation\\judgment)}] at
(-7.5,9) {C};
\node[label={[align=center]below:(man or\\ \ machine)}]  at (-1.5,9.5) {A};
\node[label={[align=center]below:\ (woman)}]  at (-1.5,7.4) {B};
\end{tikzpicture}
\caption{The schema of the original experimental arrangement where the interrogator must decide which of X and Y are A and B.\label{fig_turingtest}}
\end{figure}
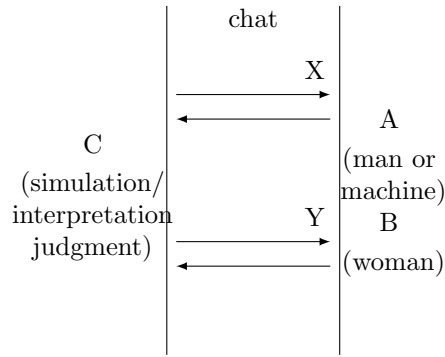

The Turing test (originally called the Imitation Game) is a special kind of conversation that was introduced by Alan Turing in 1950 \cite{turingtest}, and was intended to investigate computer programs from point of view of intelligence. The Fig. \ref{fig_turingtest} shows the original experimental arrangement used in the test. We believe that this test measures the intelligence of the person C at least as much as it measures the intelligence of X or the intelligence of Y.
Therefore, in the following, we suggest a more objective interrogator to replace the person C. 
But it is to be noted that the possibility of instrumental observation of interrogator, such as EEG for detecting N400 violations \cite{N400}, will not be looked into.

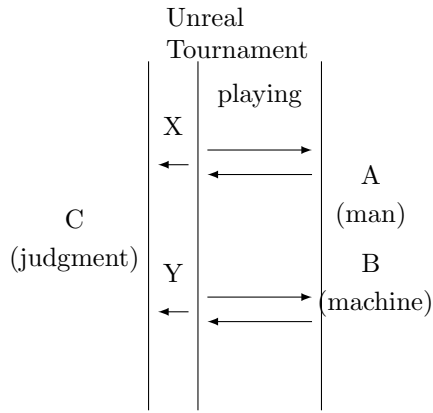
\begin{figure}[t]
\centering
\begin{tikzpicture}[ scale=0.65]
\node[label={[align=left]above:Unreal\\Tournament}]  at (-4.2,11.5) {};
\node[label={[align=center]above:playing}]  at (-3.75,10.5) {};
\node (cv2) at (-6,4.5) {};
\node (cv1) at (-6,12) {};
\draw  (cv1) edge (cv2);
\node (v2) at (-5,4.5) {};
\node (v1) at (-5,12) {};
\node (v3) at (-2.5,12) {};
\node (v4) at (-2.5,4.5) {};
\draw  (v1) edge (v2);
\draw  (v3) edge (v4);
\node (v5) at (-5,10) {};
\node (v6) at (-2.5,10) {};
\node (v12) at (-5,6.5) {};
\node (v11) at (-2.5,6.5) {};
\tikzstyle{myedgestyle} = [-latex]
\draw[-latex]  (v5) edge (v6);
\node (v8) at (-5,9.5) {};
\node (v7) at (-2.5,9.5) {};
\draw[-latex]  (v7) edge (v8);
\node at (-5.5,10.5) {X};
\node (cvx8) at (-6,9.7) {};
\node (cvx7) at (-5,9.7) {};
\draw[-latex]  (cvx7) edge (cvx8);
\node (v9) at (-5,7) {};
\node (v10) at (-2.5,7) {};
\draw[-latex]  (v9) edge (v10);
\draw[-latex]  (v11) edge (v12);
\node at (-5.5,7.5) {Y};
\node (cvx82) at (-6,6.7) {};
\node (cvx72) at (-5,6.7) {};
\draw[-latex]  (cvx72) edge (cvx82);
%\node[label={[align=center]below:Tournament}] at (-7.5,8.6) {C + Unreal};
\node[label={[align=center]below:(judgment)}] at (-7.5,8.6) {C};
\node[label={[align=center]below:(man)}]  at (-1.5,9.5) {A};
\node[label={[align=center]below:\ (machine)}]  at (-1.5,7.7) {B};
\end{tikzpicture}
\caption{Schematic representation of actors in the original 2K BotPrize Contest.\label{fig_botprize}}
\end{figure}

In the case of spoken languages, to replace the human interrogator with a software seems to be impossible for the time being of course but searching the literature \cite{cisurvey}
we can see many examples in which the role of interrogation in the Turing test has already been changed.
For example, in the BotPrize Contest \cite{2kbotp} human judges have taken over the role of interrogator and the chat is replaced with playing with the well-known first-person shooter Unreal Tournament. Fig. \ref{fig_botprize} shows a possible schematic representation of the relationship of actors in the 2K BotPrize Contest. 
It may be noted that a number of examples, such as the Turing test track of the 2012 Mario AI Championship \cite{mario1}, \cite{mario2}, the Ms. pac-man competition \cite{pacman} or the 2009 simulated car racing championship \cite{carracing} may be interpreted by a similar model.
The test of BotPrize has been further developed in \cite{2kbotp2} by its original author in order to integrate judging into the game itself. With this development the BotPrize test has been retrofitted to become an implementation of the reverse Turing test \cite{reverseTuring}, \cite{2kbotp2} but, from our point of view, this shift in direction can also be interpreted as looking for a more objective interrogator to replace the human judges.

In the beginning of the following section, we restrict ourselves to the case of Turing machines.

\subsection{Intellectual dimension}

In manuscript 
\cite{OrchMach}, the author introduced the notion of the breed and gave a family of orchestrated machine simulation algorithms for Turing machines. A breed is simply a set of Turing machines that can cooperate with each other by synchronizing (orchestrating) their execution. At each step of the operation of the orchestrated machine, one rule is non-deterministically chosen from the applicable rules of machines of the breed to be executed on the all Turing machines where it is possible. If a machine cannot execute the chosen rule then 
that machine will be deleted from the breed. Intuitively, $N$ denotes the number of steps that the breed can take and $\left \lfloor \overline{o_2} \right \rfloor$ denotes the mean of the sizes of the breed. A C++ implementation of the first orchestration algorithm (OrchMach1) can be found in the git repository at \url{https://github.com/nbatfai/orchmach} and a full detailed introduction of orchestration algorithms is described in pseudo code in \cite{OrchMach}.

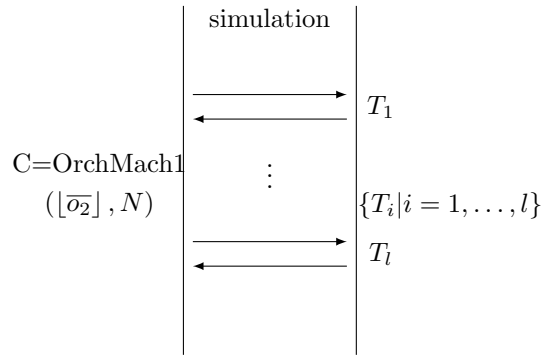
\begin{figure}[h]
\centering
\begin{tikzpicture}[ scale=0.65]
\node[label={[align=center]simulation}]  at (-4.25,11) {};
\node (v2) at (-6,4.5) {};
\node (v1) at (-6,12) {};
\node (v3) at (-2.5,12) {};
\node (v4) at (-2.5,4.5) {};
\draw  (v1) edge (v2);
\draw  (v3) edge (v4);
\node (v5) at (-6,10) {};
\node (v6) at (-2.5,10) {};
\node (v12) at (-6,6.5) {};
\node (v11) at (-2.5,6.5) {};
\tikzstyle{myedgestyle} = [-latex]
\draw[-latex]  (v5) edge (v6);
\node (v8) at (-6,9.5) {};
\node (v7) at (-2.5,9.5) {};
\draw[-latex]  (v7) edge (v8);

\node (v9) at (-6,7) {};
\node (v10) at (-2.5,7) {};
\draw[-latex]  (v9) edge (v10);
\draw[-latex]  (v11) edge (v12);

\node[label={[align=center]below:($\left\lfloor\overline{o_2} \right\rfloor, N$)}] at
(-7.7,8.6) {C=OrchMach1};
\node[label={[align=center]below:$\{T_i\vert i=1,\dots, l\}$}]  at (-0.6,8.4) {};
\node at (-4.25,8.5) {$\vdots$};
\node at (-2,9.75) {$T_1$};
\node at (-2,6.75) {$T_l$};
\end{tikzpicture}
\caption{The orchestrated machine itself has been considered as a kind of interpretation of Turing's Imitation Game. The algorithm OrchMach1 orchestrates its input Turing machines $T_1, \dots, T_l$.\label{fig_om1test}}
\end{figure}

If we turn our attention back to Turing test we can see easily that the orchestrated machine can be interpreted as a special kind of Turing test that is shown in Fig. \ref{fig_om1test}. This arrangement do not use any subjective elements and gives totally objective standard for judging the intelligence level of the investigated breed $\{T_i\}$ because the outputs of this test are the values $N$ and $\left \lfloor \overline{o_2} \right \rfloor$. 
The test itself is the simple execution of Alg. 1 defined in \cite{OrchMach}.

But how can we generalize this to the case of arbitrary computer programs? It is a real theoretical challenge because the orchestrated architecture is completely build on the Turing machine architecture now.

\begin{figure}[h]
\centering
\begin{tikzpicture}[ scale=0.65]
\node[label={[align=center]above:chess}]  at (-4.9,11.5) {};
\node[label={[align=center]above:playing}]  at (-3.75,10.5) {};
\node (cv2) at (-6,4.5) {};
\node (cv1) at (-6,12) {};
\draw  (cv1) edge (cv2);
\node (v2) at (-5,4.5) {};
\node (v1) at (-5,12) {};
\node (v3) at (-2.5,12) {};
\node (v4) at (-2.5,4.5) {};
\draw  (v1) edge (v2);
\draw  (v3) edge (v4);
\node (v5) at (-5,10) {};
\node (v6) at (-2.5,10) {};
\node (v12) at (-5,6.5) {};
\node (v11) at (-2.5,6.5) {};
\tikzstyle{myedgestyle} = [-latex]
\draw[-latex]  (v5) edge (v6);
\node (v8) at (-5,9.5) {};
\node (v7) at (-2.5,9.5) {};
\draw[-latex]  (v7) edge (v8);
\node at (-5.5,10.5) {X};
\node (cvx8) at (-6,9.7) {};
\node (cvx7) at (-5,9.7) {};
\draw[-latex]  (cvx7) edge (cvx8);
\node (v9) at (-5,7) {};
\node (v10) at (-2.5,7) {};
\draw[-latex]  (v9) edge (v10);
\draw[-latex]  (v11) edge (v12);
\node at (-5.5,7.5) {Y};
\node (cvx82) at (-6,6.7) {};
\node (cvx72) at (-5,6.7) {};
\draw[-latex]  (cvx72) edge (cvx82);
%\node[label={[align=center]below:Tournament}] at (-7.5,8.6) {C + Unreal};
\node[label={[align=center]below:(judgment)}] at (-7.5,8.6) {C};
\node
%[label={[align=center]below:(man)}] 
 at (-.6,9.6) {Garry Kasparov};
\node
%[label={[align=center]below:\ (machine)}]  
at (-.8,6.8) {Deep Blue};
\end{tikzpicture}
\caption{Could an entirely ordinary interrogator decide which of X and Y are Garry Kasparov and Deep Blue?\label{fig_pchess}}
\end{figure}
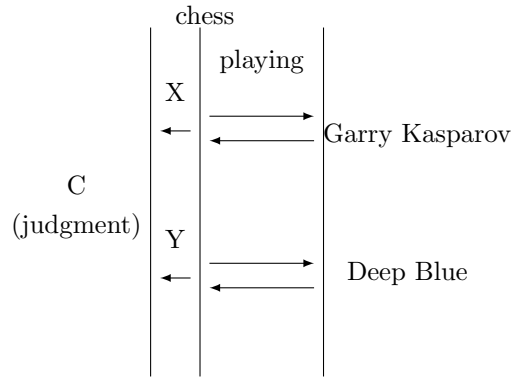

Let us consider, only as an intuitive example, the Deep Blue versus Garry Kasparov chess match (1997 rematch, Game 5, see \url{http://en.wikipedia.org/wiki/Deep_Blue_versus_Garry_Kasparov}) finished in draw after 49 moves of play. In this case, $N$ is equal to 49, $\left \lfloor \overline{o_2} \right \rfloor$ is equal to 2 so the intellectual dimension of this game equals $log_249 = 5.61471$.
Fig. \ref{fig_pchess} raises the question whether an ordinary interrogator could determine which of X and Y are Garry Kasparov and Deep Blue in Turing's Imitation Game? In this regard, it should be noted that Kasparov felt that Deep Blue had human-like behavior \cite{kasparov}. In this sense, Kasparov may qualify as an interrogator. But other conceptual problems may also be raised in relation to the test.
For example, in our opinion, if a computer program (marked by letter A in Fig. \ref{fig_turingtest}) passed the Turing test then it would take over the role of the interrogator (marked by C).  What happens if this C=A interrogator asks the question, do you have an infinite loop? With this modification, we can define the interrogation as a decision problem.

In spite of the success of developments in the field of Turing tests such as Weisenbaum's Eliza \cite{eliza} or Loebner Prize \cite{loebner} awarded computer programs it is a plain fact that Turing's prediction still has not come true. We still do not have computer programs that participate in and understand conversations as humans do. From the beginning there are opinions voiced that artificial intelligence is not going to the right direction \cite{Dreyfus1}. It is quite easy to believe that classic -- that is, written from scratch by human programmers -- computer programs such as C++, Lisp, Python, Java or Turing programs cannot pass the test. All of these programs are designed to be run within an entirely isolated environment. All of their parts are designed directly, by human experts, to fulfill their exact predetermined purpose and they do not contain any historical accidental elements. In contrast, living systems such as natural languages have no pre-planned parts and come into existence as a result of accidental processes \cite{Neumann}.
The concept of orchestrated machines lies between these two extremes using elements from both. On the one hand, the input of the orchestrated machine is a set of ordinary Turing machines. It can be seen as a designed  feature. But on the other hand the executions of input Turing machines can influence, improve or confuse the operations of each other. This operation ``simulates'' a dissipative-like behavior. It can be seen as an accidental feature.

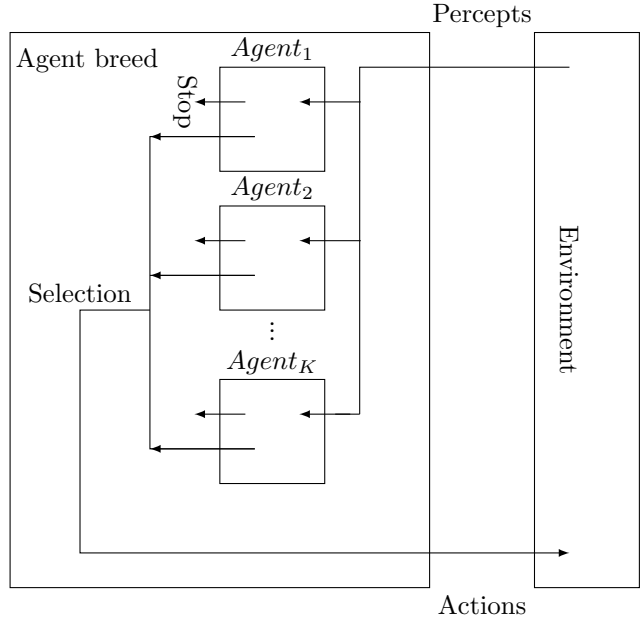
\begin{figure}[h]
\centering
\begin{tikzpicture}[ scale=0.92]
\draw  (-5.5,5.5) rectangle (0.5,-2.5);
\draw  (2,5.5) rectangle (3.5,-2.5);
\draw  (-2.5,5) rectangle (-2.5,5) node (v1) {};
\draw  (v1) rectangle (-1,3.5);
\draw  (-2.5,3) rectangle (-1,1.5);
\draw  (-2.5,0.5) rectangle (-1,-1);
\draw (2.5,5) -- (-0.5,5) -- (-0.5,0) node (v6) {};
\draw (-0.5,0) -- (-0.85,0) {};
\node (v2) at (-0.35,4.5) {};
\node (v4) at (-0.35,2.5) {};
\node (v3) at (-1.5,4.5) {};
\node (v5) at (-1.5,2.5) {};
\node (v7) at (-1.5,0) {};
\draw[-latex]  (v2) edge (v3);
\draw[-latex]  (v4) edge (v5);
\draw[-latex]  (v6) edge (v7);
\node[label=center:\rotatebox{-90}{Stop}] (v9) at (-3,4.5) {};
\node (v11) at (-3,2.5) {};
\node (v13) at (-3,0) {};
\node (v8) at (-2,4.5) {};
\node (v10) at (-2,2.5) {};
\node (v12) at (-2,0) {};
\draw[-latex]  (v8) edge (v9);
\draw[-latex]  (v10) edge (v11);
\draw[-latex]  (v12) edge (v13);
\draw (-2,4) -- (-3.5,4) -- (-3.5,-0.5) -- (-2,-0.5);
\draw (-2,2) -- (-3.5,2);
\draw[-latex] (-3.5,1.5) --(-4.5,1.5) -- (-4.5,-2) -- (2.5,-2);
\node at (-4.4,5.1) {Agent breed};
\node at (1.25,5.75) {Percepts};
\node at (1.25,-2.75) {Actions};
\node[label=below:\rotatebox{-90}{Environment}] at (2.5,3) {};

\node at (-1.75,.75) {$Agent_K$};
\node at (-1.75,3.25) {$Agent_2$};
\node at (-1.75,5.25) {$Agent_1$};
\node at (-4.5,1.75) {Selection};

\draw[-latex] (-2,4) -- (-3.5,4) ;
\draw[-latex] (-2,2) -- (-3.5,2) ;
\draw[-latex] (-2,-0.5) -- (-3.5,-0.5) ;

\node[label=center:\rotatebox{-90}{...}] at (-1.75,1.2) {};

\end{tikzpicture}
\caption{This ``orchestrated Turing test'' architecture called OrchAgent is intended to support us in searching and developing intelligent agent breeds.\label{fig_orchAgent}}
\end{figure}

Essentially the problem is how to apply the orchestrated machine algorithms such as OrchMach1 to high level programming languages. Let’s consider a set of Lisp programs. As a first step we would have come up with a new algorithm to orchestrate the Lisp programs like the OrchMach1 does with Turing machines. Then we would have to write an interpreter to implement it.  Unfortunately, the first step is not trivial. It seems easier to solve it at a higher abstraction level. Thus in this section we try to give a simple orchestration algorithm for intelligent software agents. 
Similarly to Turing machine breeds, agent breeds are simply finite sets of specialized intelligent software agents. 
This abstract machine is called \textit{agent breed} and it is shown in Fig. \ref{fig_orchAgent}, where the regular part of the terminology is taken from the book \cite{RussellNorvig}. 
Only as an intuitive example consider some conversational agents. Let $Agent_1$ be the Weisenbaum's Eliza program, and let $Agent_2$ be an AIML-based (Artificial Intelligence Markup Language, \cite{AIML}) chatbot, and so on. An agent halts if it cannot give an adequate answer to some question. For example, the original Eliza will always stop at the first step, because it only transforms its input into an answer without using any internal representation of the conversation \cite{Norvig}. The AIML-based programs will trivially stop if they interpret the pattern $<\!\!pattern\!\!>\!\!*\!\!<\!\!/pattern\!\!>$ without any $<\!\!topic\!\!>$ or $<\!\!that\!\!>$ elements.
 
The concept of agent breeds shown in Fig. \ref{fig_orchAgent} allows us to automatically develop the intelligent agents. 
Simply we can search for such type of agent breeds that can maximize $log_KN$, where $N$ equals the number of percept-action pairs. Here we may note that Winograd Schema Challenge-based tests \cite{ws1}, \cite{ws2} may allow us to automate the percepts.
At this moment we are working on the high level design of a project, called ``Man's best friend''. 
This software project will be based on the agent breed concept. It is introduced in the Sect. \ref{carHLAI}.
For Turing machine breeds a similar research project has already been initiated, it is presented in Sect. \ref{ref_game}.

In our previous paper \cite{OrchMach}, we raised the question of whether there is an interesting relation between the two intelligence functions IQ and EQ.  The empirical estimates of these functions shown in Fig. 5.  of \cite{OrchMach} suggests that IQ and EQ functions may follow a Pareto distribution or power law:
\[
IQ(\left \lfloor \overline{o_2} \right \rfloor) \sim \left(EQ(N)\right)^D
\]
and using Theorem 5 of  \cite{OrchMach} it follows that
\begin{align*}
N \le IQ(\left \lfloor \overline{o_2} \right \rfloor) &\sim \left(EQ(N)\right)^D\\
\left \lfloor \overline{o_2} \right \rfloor \le EQ(N) &\sim \left(IQ(\left \lfloor \overline{o_2} \right \rfloor)\right)^{-D}.
\end{align*}
This is of course only a conjecture. The quantities $log_{\left \lfloor \overline{o_2} \right \rfloor}N$
and 
$log_{EQ}IQ$ 
are called intellectual dimension.
A chess example for this definition has already been given earlier in this section. As another intuitive example, now consider Penrose and Hameroff's Orch OR model of consciousness \cite{OrchOR}. According to this model, a human brain has 40 conscious moments per second. These moments are produced by roughly 20.000 neurons  \cite[p. 61]{OrchOR}. So we may write that the intellectual dimension of this neural network is equal to $log_{2.304*10^6} 20000 =1.479293$ and the intellectual dimension of the corresponding microtubular network that is contained by this neural network is equal to $log_{2.304*10^6} (2*10^{11})=0.5630002$, where $2.304*10^6$ is the number of moments in the 16 hours of awake in a day.
Similar speculative intellectual dimension computations will be introduced in the next section for robocars.

\subsection{Man's best friend\label{carHLAI}}

But why would we look for intelligent agent breed shown in Fig. \ref{fig_orchAgent}? We need a challenge to motivate us.
The spread of driverless robocars will be a significant milestone in the development of artificial intelligence.
Several new notions, such as robotaxi and sleeper car, were born in this research and development area \cite{robocar}.
The explosive development of driverless cars invites the obvious question: will it be possible to replace the enjoyment of driving a human-driven car? Actually, we think that this question to be answered is easier than we expected.  Look at the parking of a shopping center on a Friday afternoon, parked cars fill the spaces wherever we look. Very soon now these cars will be equipped with the very latest sensors. At present, cars have ``natural'' actuators such as motors and wheels and will have considerable computing capability. 
We believe all of these will offer the ideal environment to develop a Human-Level AI and will provide a renewed impetus for the study of Turing test. By our approach the first mass produced driverless cars must have roughly the same intelligence as dogs. In this interpretation, these smart cars will become man's best friend in the near future. 
In the following the conversation-based user interfaces of cars are referred to as ``Man's best friend''-like systems. 

In the case of the search for Turing machine breeds we have already initiated a similarly structured project called  GoDNGoD, but while it is based on a game, the development of the ``Man's best friend'' should be envisioned as an open source industrial project. Our immediate research goal in this project is to create a framework in which we can automatically search for such agent breeds that will be able to orchestrate conversational Lisp agents to maximize their intellectual dimension.
It should be noted here that percepts of the architecture in Fig. \ref{fig_orchAgent} can be supported by Winograd Schema Challenge-based tests \cite{ws1}. 
In another attempt to create a ``Man's best friend''-type chat system \cite{SAMU} we try to use classical techniques such as Q-learning. In either case, we plan to introduce some kind of orchestrated machines and intellectual dimension based variant of the Imitation Game for benchmarking of the chat systems to be developed.

\section{GoDNGoD\label{ref_game}}

Many people collect specimens of insect or butterfly species.
The game GoDNGoD gives the opportunity to collect Turing machine breeds, in such sense that they can be regarded as digital organisms. For example, a contained Turing machine may be thought of as an organelle.
Two sample specimens of Turing machine breeds called ``Breedi Turingi'' and ``Turingus Tri Breedus'' are shown in Fig. \ref{fig_godngod}.

The name GoDNGoD stands for a recursive acronym for ``\underline{GoD}NGoD is \underline{N}ot \underline{GoD}'', where GoD = Game of Darwin.

\begin{figure*}[h]
\centering
\includegraphics[scale=.6]{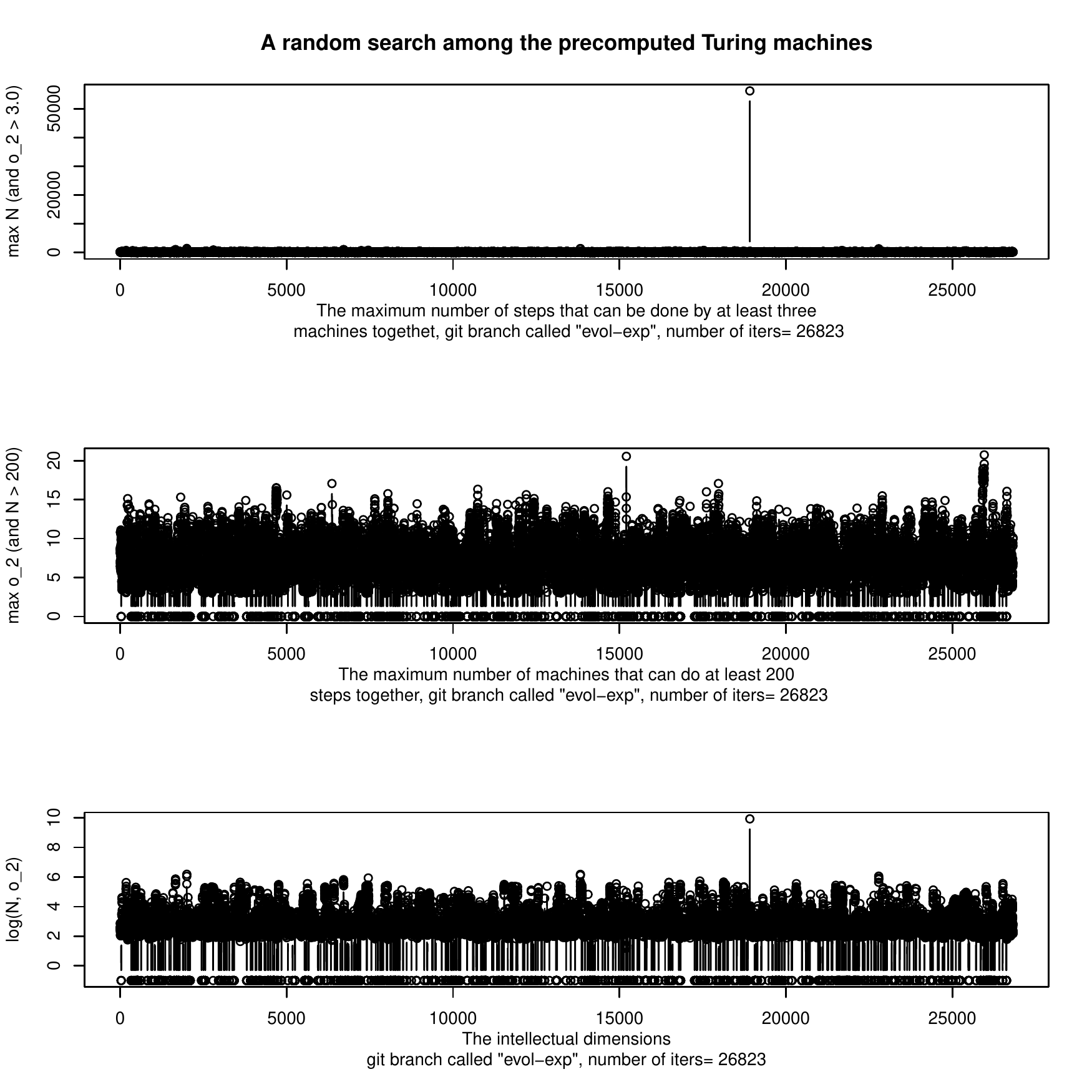}
\caption{These are R plots of an experiment that found a 10-dimensional breed shown in right side of Fig. \ref{fig_godngod} in detail.\label{fig_50}}
\end{figure*}

The game consists of two parts. The one part is a volunteer computing system that uses computational capabilities of volunteers to simulate (orchestrate) Turing machine breeds. It is based on the project \url{https://github.com/nbatfai/orchmach} that has been further developed to search automatically for breeds that promise to be interesting. We use simple evolutionary-like mechanism for searching among some well-known (such as Marxen and Buntrock's Busy Beaver champion machines \cite{MarxenBuntrock}, Uhing's machines or Schult's machines \cite{MichelSurvey}) and 20.000 other halting Turing machines. We search for such type of breeds that can maximize their intellectual dimension. The exact list of the machines and the precise description of the searching process can be found in the source files. For example, the R plots of an experiment that we got from using the code \url{https://github.com/nbatfai/orchmach/blob/master/orchmach1-exp2-evol.cpp} can be seen in Fig \ref{fig_50}. This experiment found  a 10-dimensional breed shown in right side of Fig. \ref{fig_godngod} in detail.

\begin{figure*}[h]
\centering
\includegraphics[scale=.28]{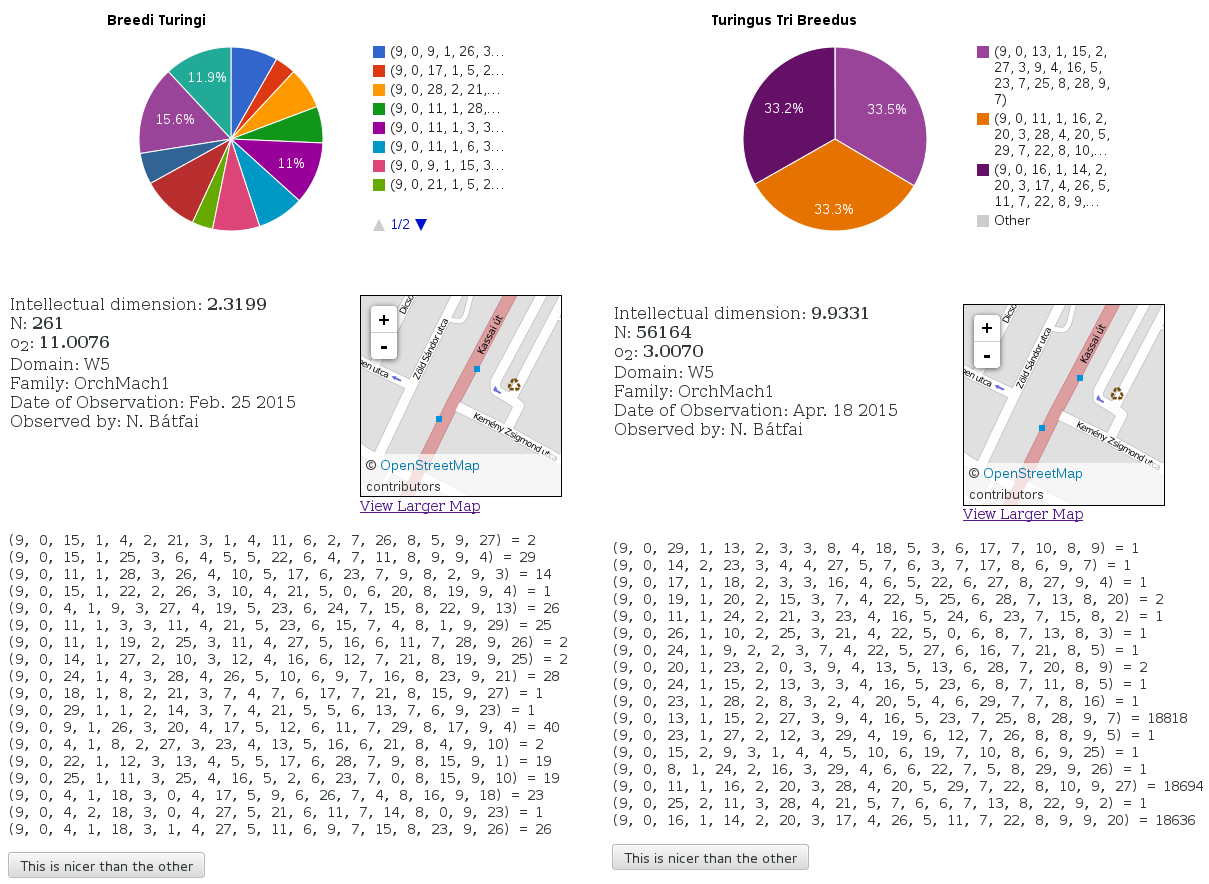}
\caption{A first draft of the GoDNGoD's user interface elements.  A panel contains 
a fancy name,  basic information such as intellectual dimension, $N$ and $o_2$, the name of the observer, the included Turing machines with the notations of \cite{running_time_arxiv} (here there is a number after the equal sign that shows how many times the given machine was selected for execution)  
and finally an OpenStreetMap snippet of latitude and longitude where the breed was found. (Pie charts are created with the Google Visualization API.)
\label{fig_godngod}}
\end{figure*}

The other one is an administrative and visualization part. It is built on the top of the previous layer. The main task of it is to record the breeds and especially to delete breeds found to be infinite. 
The visualization interface shown in Fig. \ref{fig_godngod}. Our immediate research goal in this game is to use an artificial neural network for performing Turing breeds classification automatically. This layer has not been available yet for public use.

\section{Conclusion}

In the near future we are going to develop ``Man's best friend''-like systems that will be based on some orchestration architecture. A benefit, and goal, of this  paper is provide a link between the theoretical background of orchestrated machines \cite{OrchMach} and the practical applications. 
To support it, we have introduced the artificial life game GoDNGoD.
In addition, as a possible ``killer application'' for artificial intelligence we have described and brainstormed a vision of whether it will be possible to replace the enjoyment of driving a human-driven car.
We have introduced a fractal-like measure called intellectual dimension that can intuitively examine the cooperation of conversational agents. We are going to compute and examine variants of it for benchmarking of the ``Man's best friend''-like systems to be developed.

\section*{Acknowledgment}

The computations used in the preliminary experiments of this paper were partially performed on the NIIF High Performance Computing supercomputer at University of Debrecen.

\bibliography{orchmachgame}

\end{document}